# Responsible AI in Construction Safety: Systematic Evaluation of Large Language Models and Prompt Engineering

Farouq Sammour [1]; Jia Xu [2]; Xi Wang, Ph.D., A.M.ASCE[3]; Mo Hu, Ph.D., A.M.ASCE [4]; and Zhenyu Zhang, Ph.D.[5]

[1]Ph.D. student, Dept. of Construction Science, Texas A&M Univ., College Station, TX 77840. Email: farouq@tamu.edu

[2]Ph.D. student, Dept. of Construction Science, Texas A&M Univ., College Station, TX 77840. Email: jiaxu@tamu.edu

[3]Assistant Assistant Professor, Dept. of Construction Science, Texas A&M Univ., College Station, TX 77840. Email: xiwang@tamu.edu

[4]Assistant Professor, Dept. of Construction Science, Texas A&M Univ., College Station, TX 77840. Email: mohu@tamu.edu

[5]Assistant Professor, Dept. of Construction Science, Texas A&M Univ., College Station, TX 77840 corresponding author). Email: z.zhang@tamu.edu

**ABSTRACT:**

Construction remains one of the most hazardous sectors. Recent advancements in AI, particularly Large Language Models (LLMs), offer promising opportunities for enhancing workplace safety. However, responsible integration of LLMs requires systematic evaluation, as deploying them without understanding their capabilities and limitations risks generating inaccurate information, fostering misplaced confidence, and compromising worker safety. This study evaluates the performance of two widely used LLMs, GPT-3.5 and GPT-4o, across three standardized exams administered by the Board of Certified Safety Professionals (BCSP). Using 385 questions spanning seven safety knowledge areas, the study analyzes the models' accuracy, consistency, and reliability. Results show that both models consistently exceed the BCSP benchmark, with GPT-4o achieving an accuracy rate of 84.6% and GPT-3.5 reaching 73.8%. Both models demonstrate strengths in safety management systems and hazard identification and control, but exhibit weaknesses in science, mathematics, emergency response, and fire prevention. An error analysis identifies four primary limitations affecting LLM performance: lack of knowledge, reasoning flaws, memory issues, and calculation errors. Our study also highlights the impact of prompt engineering strategies, with variations in accuracy reaching 13.5% for GPT-3.5 and 7.9% for GPT-4o. However, no single prompt configuration proves universally effective. This research advances knowledge in three ways: by identifying areas where LLMs can support safety practices and where human oversight remains essential, by offering practical insights into improving LLM implementation through prompt engineering, and by providing evidence-based direction for future research and development. These contributions support the responsible integration of AI in construction safety management toward achieving zero injuries.

**INTRODUCTION:**

The construction industry remains the most hazardous sector in the United States, accounting for nearly 20% of workplace fatalities (BLS 2023). While experienced safety professionals are crucial for injury prevention, a pervasive shortage in safety management roles, especially in small businesses, has limited companies' capacity to implement comprehensive safety programs (Al-Bayati 2021; Ozmec et al. 2015). This challenge is further intensified by the increasingly complex construction project demands (Zhou et al. 2015) and cultural and language barriers posed by a growing migrant workforce (Vignoli et al. 2021). In response, safety expertise driven by Artificial Intelligence (AI) has gained attention as a potential remedy (Bigham et al. 2018). Emerging technologies, particularly large language models (LLMs), offer unprecedented opportunities to meet this need. The release of OpenAI's GPT-3.5 sparked widespread interest in LLMs, showcasing their ability to process human input and generate contextual, coherent responses (Saka et al. 2024). This breakthrough has paved the way for exploring the potential of LLMs in construction safety practices to reduce workplace injuries.

Applications of LLMs in construction safety are now beginning to emerge (Charalampidou et al. 2024). For instance, LLMs have been integrated into virtual reality platforms as training agents, providing safety information to migrant construction workers in their native language (Hussain et al. 2024). Another study explored LLMs' potential to visually identify safety hazards in images of construction sites for real-time detection of on-site risks (Samsami 2024). Additionally, Smetana et al. (2024) utilized LLMs to identify accident causes by analyzing textual data on highway construction accidents, surpassing traditional statistical methods that primarily focus on numerical data. These examples highlight the potential of LLMs to enhance various aspects of construction safety.

However, despite their potential, levering LLMs for safety management requires a careful evaluation of their subject-specific knowledge. Studies in other fields have revealed that LLMs can generate biased and inaccurate content when trained on skewed data (Schramowski et al. 2022). LLMs are also prone to “hallucination”, fabricating outputs not grounded in factual data (Liu et al. 2024). Deploying LLMs without understanding their strengths and limitations poses significant risks. To address these risks, it is crucial to implement well-considered designed countermeasures, such as prompt engineering and model fine-tuning. Prompt engineering involves crafting clear instructions and contexts to guide LLMs in generating accurate outputs (Chen et al. 2023a). For example, employing a Chain-of-Thought (CoT) prompting strategy can improve an LLM's step-by-step reasoning. This approach effectively leverages the model's pre-trained mathematical knowledge, leading to more accurate solutions in mathematical problem-solving (Pham et al. 2024). However, there are instances where an LLM's pre-trained knowledge may not suffice for specific tasks or domains. In such cases, model fine-tuning becomes essential. A prime example of this is Google's Med-PaLM 2, which was fine-tuned on medical data to enhance the model's clinical language understanding (Qian et al. 2024). This specialized fine-tuning allowed the model to extend its capabilities beyond its original general-purpose design.

Responsible deployment of LLMs in construction safety management begins with a systematic evaluation of their capabilities and limitations within the specific context of the field. Currently, there is a noticeable gap in such evaluations. This contrasts with other fields, such as law, medicine, education, and computer science, where extensive evaluations have been conducted to illuminate responsible ways for professionals to utilize LLMs (Bommarito et al. 2023; Callanan et al. 2023; Stribling et al. 2024). While these evaluations in other domains offer some insights, they cannot substitute for a field-specific assessment of LLMs in construction safety. This is because the performance of LLMs can vary significantly across different fields due to the varying amounts and types of pre-trained data available.

This research addresses a critical gap by evaluating LLM capabilities and limitations using standardized safety certification exams. The study makes three key contributions: First, it examines LLMs' knowledge across seven safety knowledge areas (KAs), providing insights into the areas where these AI models are reliable and where human oversight is necessary. Second, the study rigorously examines prompt-related factors to determine their impact on LLM performance. These findings enable construction professionals to optimize their use of LLMs in safety applications through proper prompting techniques. Finally, we uncover inherent knowledge gaps and limitations in LLMs to inform the future design of advanced countermeasures. Altogether, this study lays the groundwork for the responsible and effective integration of LLMs in construction safety management, ultimately contributing to safer workplaces.

## LITERATURE REVIEW:

### *Large Language Models (LLMs)*

LLMs have emerged as a transformative technology in the field of Artificial Intelligence (AI). These models are pre-trained on vast text datasets using natural language processing and deep learning techniques. A key factor influencing the performance of LLMs is the number of parameters they contain, with larger models generally capturing more knowledge and complex language patterns. For instance, OpenAI's GPT-3.5 boasts 175 billion parameters, while GPT-4o is estimated to have an impressive 1.96 trillion parameters (OpenAI 2024). However, the increased capabilities of larger models come at the cost of greater computational demands, resulting in slower response time and higher operational expenses (Chen et al. 2023b). This trade-off requires careful consideration when selecting models for specific applications. To accommodate diverse use cases, many LLM providers offer both large and smaller parameter versions (Brin et al. 2023). Despite these advancements, a significant research gap remains in comprehensively assessing the subject-specific expertise of LLMs in the context of construction safety management, regardless of model size.

### *Evaluation of LLMs*

Standardized tests, such as professional certification exams, have emerged as a reliable method for evaluating LLMs' domain expertise (Katz et al. 2024; Long et al. 2023). This approach offers a structured and objective assessment, allowing for direct comparisons with human performance on established benchmarks (Sallam et al. 2024). For example, Gilson et al. (2023) utilized the United States Medical Licensing Examination (USMLE) to assess LLMs' medical knowledge, finding it comparable to that of an average third-year medical student. Existing evaluation studies primarily focus on accuracy by prompting each question only once, but they often overlook two critical aspects of LLM performance: consistency and reliability. LLMs are known for their randomness in responses, yet it remains largely unclear how consistently they can provide the same answer to a question when prompted multiple times. Also, the reliability of LLMs' performance across multiple exam attempts is not well understood. These factors are crucial in determining the trustworthiness of LLMs as tools in professional domains (Lin et al. 2024). To our knowledge, Wang et al. (2023) is the only paper that investigated this issue. They found that LLM performance varies when answering medical Multiple-Choice Questions (MCQs), yet this concern remains unknown in construction safety.

***Prompting Techniques***

The domain expertise of LLMs is significantly influenced by their pre-trained knowledge. However, strategic use of prompts can optimize LLM outputs by providing necessary context and instructions (Marvin et al. 2024). Various prompting techniques have been developed to enhance model performance. Direct prompting (DP) involves giving straightforward instructions, such as "Translate the following sentence to French". The Chain-of-Thought (CoT) strategy encourages LLMs to engage in step-by-step reasoning, and Chen et al. (2023a) demonstrated that adding the instruction "Let's think step-by-step" dramatically improved GPT-3's accuracy on mathematical tasks from 17.7% to 78.7%. Moreover, Few-Shot (FS) prompting supplies examples to illustrate ideal outputs (Tang et al. 2024). While these prompting techniques generally enhance task performance, their effectiveness can vary across different domains and tasks (Orlanski 2022). This variability underscores the importance of thoughtful prompt design and context-specific evaluation to maximize model performance. In construction safety management, research on prompting techniques is still in its early stages. Only one study by Samsami (2024) has explored this area, demonstrating that FS prompts outperform DP prompts in vision-based hazard identification tasks. A comprehensive investigation of prompting techniques for text-based safety tasks is urgently needed to advance this field.

***Output Requirements***

Beyond prompt techniques, output requirements also play a crucial role in human-LLM interaction. Structured generation, which involves producing content in standardized formats such as JavaScript Object Notation (JSON), ensures that users can extract AI-generated information in a consistent and machine-readable fashion (Patiny and Godin 2023). However, evidence suggests that format requirements can impact LLM performance. Chu et al. (2024) found that instructing LLMs to include both reasoning and answers in their output, with reasoning preceding the answer, often leads to improved accuracy. This suggests that users should avoid requesting immediate answers without explanations. LLMs also exhibit position bias when handling MCQs by favoring options presented first (Li et al. 2024). To address this, Zheng et al. (2023) proposed a label-free approach, in which LLMs are prompted to spell out the correct option without using traditional labels like A, B, C, or D. However, Rui Tam et al. (2024) discovered that the influence of format requirements can vary across specific prompts, models, and tasks. The complex interplay between structured output requirements and LLM performance, particularly within construction safety contexts, requires further investigation.

In summary, while LLMs demonstrate considerable potential in construction safety practices, significant research gaps remain. Further studies are required to evaluate existing models' accuracy, consistency, and reliability in standardized safety exams. There is also an urgent need to explore effective prompting techniques and assess the impact of structured output requirements, as these elements form critical yet underexplored communication protocols between users and LLMs. The goal of this study is to address these knowledge gaps to enable construction professionals to utilize LLMs responsibly and effectively.

## METHODS

### *Experimental Setup*

#### Dataset

This study utilized a dataset of 385 MCQs from three certification exams administered by the Board of Certified Safety Professionals (BCSP): Associate Safety Professional (ASP), Certified Safety Professional (CSP), and Construction Health and Safety Technician (CHST). ASP and CSP are industry-recognized exams in the U.S. for evaluating technical expertise and managerial knowledge in occupational safety and health, while CHST focuses specifically on construction safety practices. To ensure the questions were not part of any pre-trained data in LLMs, the MCQs were sourced from two proprietary, non-public databases: the official BCSP test bank and the Pocket Prep test bank. Pocket Prep, a widely used exam preparation application, provides practice questions adapted from the "Safety Professional's Reference and Study Guide " (Yates 2020). The MCQs were evenly distributed across seven core safety KAs, as detailed in Table 1, with 55 questions allocated per KA. Each question consisted of four answer options, with only one correct answer.

Table 1: Detailed Overview of Safety Knowledge Areas

| Knowledge Area | Knowledge of |
| --- | --- |
| 1. Science and Math | • Concepts in physics, chemistry, anatomy, physiology, and mathematics relevant to safety<br>• Statistical methods used to interpret safety data<br>• Calculations for safety-related tasks, such as containment volumes |
| 2. Management Systems | • Components of safety management systems<br>• Steps involved in conducting effective incident investigations<br>• Methods for budgeting and allocating resources in safety initiatives<br>• Approaches to improve organizational safety culture |
| 3. Safety Hazard Identification and Control | • Common workplace safety hazards<br>• The hierarchy of controls and how to apply it effectively<br>• Risk assessment methods to evaluate hazard severity and probability<br>• Monitoring equipment and corrective actions for hazard control |
| 4. Ergonomics, Industrial Hygiene, and Occupational Health | • Ergonomic risks related to workplace design and task layout<br>• Common occupational health hazards<br>• Control measures to mitigate health risks |
| 5. Emergency Response and Fire Prevention | • Components of an emergency response plan<br>• Fire prevention and suppression measures relevant to workplace risks<br>• Roles and responsibilities in crisis management for natural disasters, chemical spills, or violent incidents<br>• First aid and emergency medical measures for workplace readiness |
| 6. Training, Education, Communication, and Leadership | • Adult learning principles<br>• Methods for assessing training needs and competencies<br>• Effective safety communication strategies<br>• Leadership skills that foster proactive safety behavior |
| 7. Law, Ethics, and Risk Management | • Legal requirements and standards for workplace safety<br>• Ethical principles and the BCSP Code of Ethics<br>• Guidelines for confidentiality, liability, and employee rights<br>• Risk management strategies about legal and ethical obligations |

**Test Environment**

A Python-based test environment was developed for this study, utilizing the OpenAI Application Programming Interface (API) to access the GPT-3.5 and GPT-4o models. Figure 1 provides a sample configuration of the prompt given to the GPT models in this study. Specifically, the system prompt defined the model's role, incorporating specific prompt techniques and the desired JSON output structure. MCQs were presented sequentially, one at a time, in the user prompt. To mitigate the "lost-in-the-middle" problem—where LLMs may struggle with extended input by focusing disproportionately on the beginning and end of the context (Hsieh et al. 2024)—the model's memory was reset between each MCQ. Given the fact-based and non-creative nature of MCQs, the temperature for both GPT models was set to 0, minimizing output randomness. We fixed the seed number to 42 for result reproducibility.

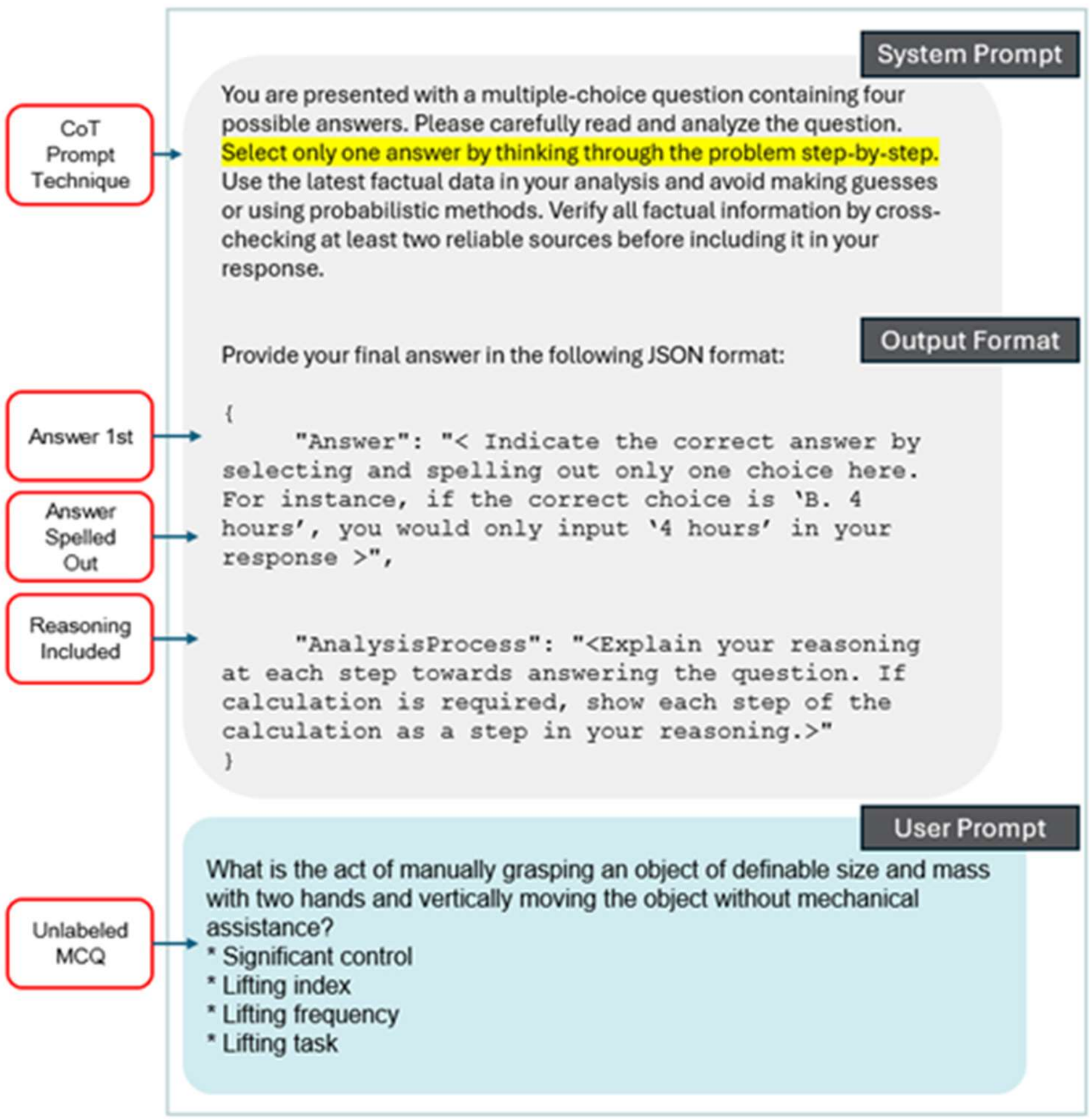

Figure 1: Example prompt structure demonstrating the CoT technique with JSON-constrained output. The answer is explicitly spelled out first, followed by the reasoning. The user prompt presents questions with unlabeled options.

**Experimental Factors**

This study examined three promoting techniques within the safety KAs: Direct Prompting (DP), Chain of Thought (CoT), and Few-Shot (FS). These techniques were carefully selected to represent a spectrum of complexity and strategies for eliciting responses from LLMs. DP is used as the baseline technique due to its simplicity in basic LLM interactions. CoT is anticipated to increase accuracy in LLM responses by enabling a multi-step, detailed reasoning process. An additional improvement is expected with FS prompting, where seven examples of desirable CoT processes are provided to guide the model's reasoning. In addition, the study focuses on three factors related to output structure: reasoning inclusion, reasoning order, and answer format. These factors were selected for their practical relevance to LLM users, who must decide, during structured generation, whether to explicitly include reasoning in the output, whether to present reasoning before the answer, and the desired format for answer delivery (i.e., labeled or unlabeled answer choices for MCQs). The first two factors have been studied previously but without conclusive evidence (Chu et al. 2024; Li et al. 2024). The third factor, to the best of our knowledge, has not yet been explored. This study evaluated the prompts and output factors using two GPT models: GPT-3.5 and GPT-4o. It was anticipated that GPT-3.5, being a smaller model with less pre-trained knowledge, may show greater improvement when optimized with carefully designed prompts and output formats, compared to the larger GPT-4o model.

To systematically manage and identify each configuration, a naming convention denoted as "MxTxOx$_1$st_x_x" was employed. In this context, a "configuration" refers to a specific combination of model type, prompt technique, and output format settings used during testing. For example, the configuration "M4TCoTOR$_1$st_Spelled_Inc" involves the GPT-4o model (M4), utilizing the CoT prompting technique (TCoT), where reasoning is presented first (R$_1$st), answer choices are unlabeled and fully spelled out (Spelled), and reasoning is included (Inc).

**Evaluation Metrics**

This study focused on three dependent variables: accuracy, reliability, and consistency. Accuracy is defined as the percentage of correct answers provided by the LLMs, calculated by comparing the model's selected answers to the correct answers in the dataset. Reliability is defined as the model's ability to maintain similar accuracy levels across multiple exam attempts. To evaluate reliability, we maintained the same levels of independent variables, and the exams were attempted five times (Wang et al. 2024). A repeated-measures Analysis of Variance (ANOVA) was then employed to detect significant differences in average accuracy between these attempts. Finally, consistency was assessed by determining whether the LLMs provided the same responses when prompted with the same question repeatedly. To quantify the consistency, we used an entropy-based metric (Equation 1):

$$H(X) = -\sum_{i=1}^{n} p(x_i) \log_2 p(x_i) \quad (1)$$

In this equation, $H(X)$ represents the entropy of the model's responses, $p(x_i)$ is the probability of selecting a specific answer $x_i$, and $n$ is the number of possible answer choices. Given that the possible answers for the LLMs to answer MCQs are five—A, B, C, D, and Error—there are five possible responses, so $n = 5$. The range of $H(X)$ is from 0 to 2.32, where 0 indicates maximum consistency (i.e., the model always selects the same answer), and 2.32 reflects maximum inconsistency (i.e., the model selects all five answers with equal probability).

***Experimental Design and Data Analysis***

**Experiment 1: Accuracy and Error Analysis**

Experiment 1 was designed to evaluate the impact of different configurations on the accuracy of LLMs. The study employed a full factorial design with five factors, resulting in 48 unique configurations applied across seven KAs. The analysis began with descriptive statistics to provide an overview of model performance, followed by a one-way ANOVA to identify the most influential model, prompt, and output-related factors on accuracy. To gain deeper insights into model weaknesses, an error analysis was conducted on the questions with the lowest accuracy scores. We selected the three lowest accuracy questions from each KA, resulting in a total of 21 questions. For each question, two of the five model outputs were randomly chosen for detailed examination, generating a total of 42 outputs for analysis. The errors identified in these outputs were categorized into four types, following established frameworks in existing research: (1) lack of knowledge, (2) reasoning flaws, (3) memory issues, and (4) calculation errors (Bilbao et al. 2023; Jin et al. 2024; Ray 2023). This classification was validated through a structured, three-round review process involving two safety experts with relevant credentials and experience in the field.

**Experiment 2: Reliability and Consistency**

Experiment 2 focused on evaluating the reliability and consistency of the LLMs' performance across multiple trials. Sixteen configurations were randomly selected for this phase using a fractional factorial design, each tested five times (Gunst and Mason 2009; Wang et al. 2024). This approach ensured diverse experimental representation and scientific rigor while minimizing the required number of trials. To assess the reliability of LLMs, we conducted a repeated-measures ANOVA to analyze if there is significant difference in accuracy across five exam attempts. Our analysis employed a General Linear Model procedure, specifically a within-subjects design, to investigate the effect of multiple conditions (the five trials) on accuracy. This approach was chosen because it effectively handles the dependent nature of repeated measurements from the same subjects or, in this case, the same LLM configurations. Before interpreting the results, we addressed the key assumptions of repeated measures ANOVA. One critical assumption is sphericity, which requires equal variance of the differences between all pairs of conditions. To test this assumption, we performed Mauchly's Test of Sphericity. The data violated this assumption, necessitating a correction to ensure the validity of our results. Consequently, the Greenhouse-Geisser corrections were applied to adjust the degrees of freedom and p-value.

To assess the consistency of LLMs, we introduced an entropy-based metric. This metric measured the variability in model responses to identical questions over five trials. We applied this method using the same 16 configurations as in the reliability assessment. For each question, we calculated entropy values ranging from 0 to 2.32 bits. Afterward, the entropy values were averaged across the 16 configurations and standardized to ensure comparability. To gain deeper insights, we grouped the standardized entropy values into 10% accuracy bins. This step allowed us to analyze how the consistency of the models changed relative to their accuracy.

## RESULTS

### *Experiment 1: Accuracy and Error Analysis*

#### Accuracy

The accuracy range, mean, and median of GPT-3.5 and GPT-4o, averaged across seven KAs, is presented in Figure 2. Both LLMs meet the BCSP passing criteria of 60%. GPT-4o significantly (p<0.01**, see ANOVA results in Table 2) outperformed GPT-3.5, with an average accuracy of 84.6% compared to GPT-3.5's 73.8% (Figure 2). Furthermore, GPT-4o's accuracy ranged from 81% to 88.9%, with a narrower range of 7.9%. GPT-3.5 improved from 64.1% in its lowest-performing configuration to 77.6% in its best (Table 3). These values suggest that GPT-3.5 has more room for improvement when optimizing prompt and output factors. Figure 3 further illustrates this through box plots, which compare the models' accuracy across the KAs, with GPT-4o showing higher accuracy and a smaller range.

Table 2: ANOVA Results for Factors Affecting Overall Model Accuracy

| | **Type III Sum of Squares** | **df** | **Mean Square** | **F** | **Sig.** |
|---|---|---|---|---|---|
| GPT Model | 0.936 | 1 | 0.936 | 317 | 0.000 ** |
| Prompt Technique | 0.005 | 2 | 0.003 | 0.892 | 0.411 |
| Reasoning Order | 0.008 | 1 | 0.008 | 2.748 | 0.098 |
| Answer Format | 0.010 | 1 | 0.010 | 3.254 | 0.072 |
| Reasoning Inclusion | 0.004 | 1 | 0.004 | 1.265 | 0.262 |
| Knowledge Area | 1.418 | 6 | 0.236 | 80.069 | 0.000 ** |

Note: $p < 0.05$*, $p < 0.01$**

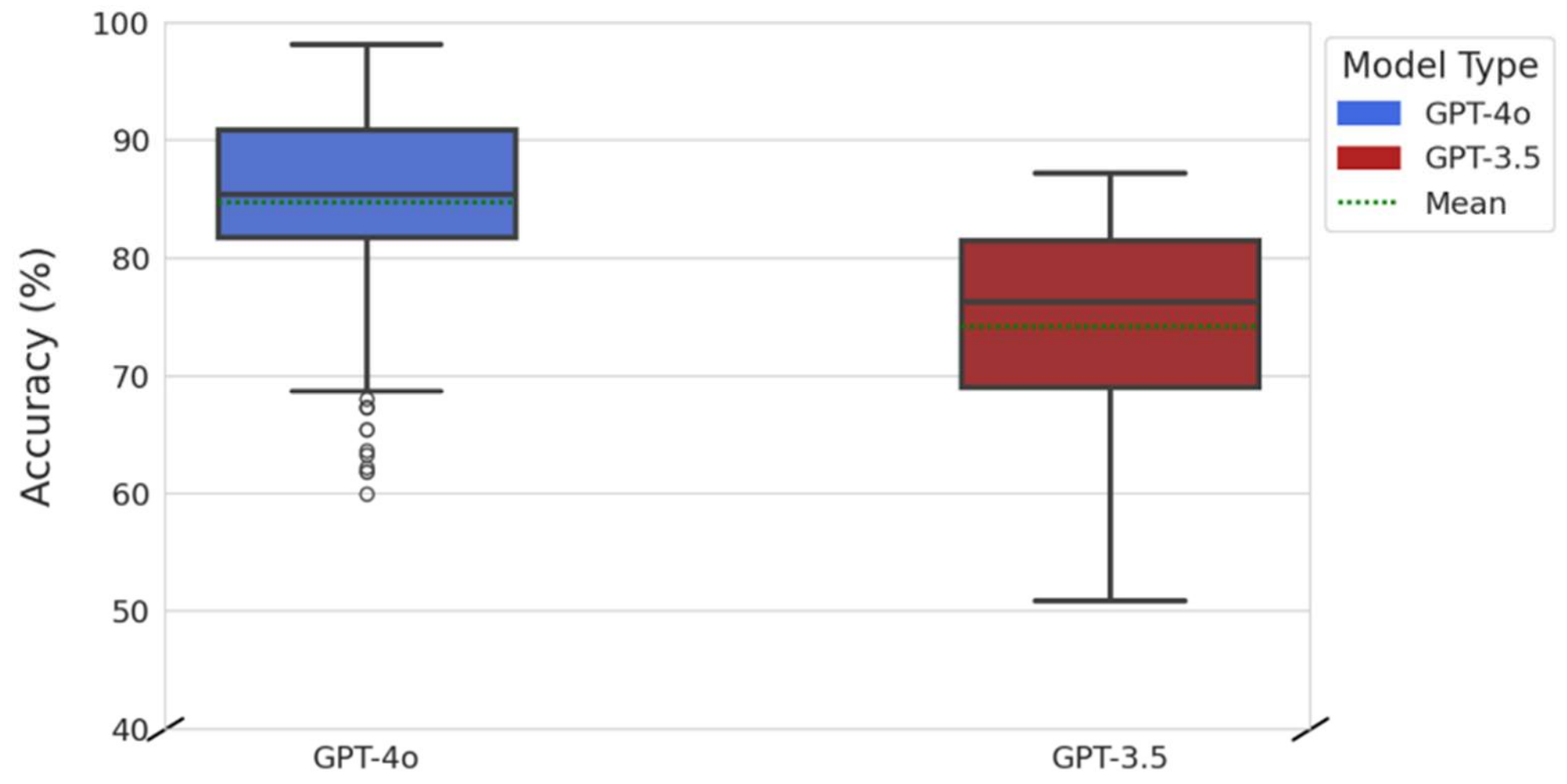

Figure 2: Comparison of accuracy between GPT-4o and GPT-3.5

Table 3: Accuracy for Top-Performing and Lowest-Performing Configurations

| | GPT-4o | | | GPT-3.5 | |
|---|---|---|---|---|---|
| **Rank** | **Configuration** | **Accuracy** | **Rank** | **Configuration** | **Accuracy** |
| 1 | $M_4T_{CoT}O_{R_1st_ABCD_Inc}$ | 88.83% | 1 | $M_3T_{FS}O_{R_1st_ABCD_Inc}$ | 77.56% |
| 2 | $M_4T_{FS}O_{A_1st_Spelled_Inc}$ | 88.31% | 2 | $M_3T_{CoT}O_{R_1st_ABCD_Inc}$ | 77.14% |
| 3 | $M_4T_{FS}O_{R_1st_Spelled_Inc}$ | 87.79% | 3 | $M_3T_{FS}O_{R_1st_Spelled_Exc}$ | 76.88% |
| 4 | $M_4T_{FS}O_{A_1st_Spelled_Exc}$ | 87.53% | 4 | $M_3T_{FS}O_{A_1st_Spelled_Exc}$ | 76.62% |
| 5 | $M_4T_{FS}O_{R_1st_Spelled_Exc}$ | 87.43% | 5 | $M_3T_{FS}O_{R_1st_Spelled_Inc}$ | 76.36% |
| … | … | … | … | … | … |
| 44 | $M_4T_{FS}O_{A_1st_ABCD_Exc}$ | 81.56% | 44 | $M_3T_{CoT}O_{A_1st_Spelled_Inc}$ | 72.47% |
| 45 | $M_4T_{DP}O_{R_1st_ABCD_Exc}$ | 81.30% | 45 | $M_3T_{Dir}O_{R_1st_Spelled_Exc}$ | 72.47% |
| 46 | $M_4T_{CoT}O_{R_1st_ABCD_Exc}$ | 81.19% | 46 | $M_3T_{FS}O_{R_1st_ABCD_Exc}$ | 71.95% |
| 47 | $M_4T_{CoT}O_{A_1st_ABCD_Exc}$ | 81.14% | 47 | $M_3T_{CoT}O_{R_1st_ABCD_Exc}$ | 71.69% |
| 48 | $M_4T_{CoT}O_{A_1st_ABCD_Inc}$ | 81.04% | 48 | $M_3T_{FS}O_{A_1st_Spelled_Inc}$ | 64.16% |

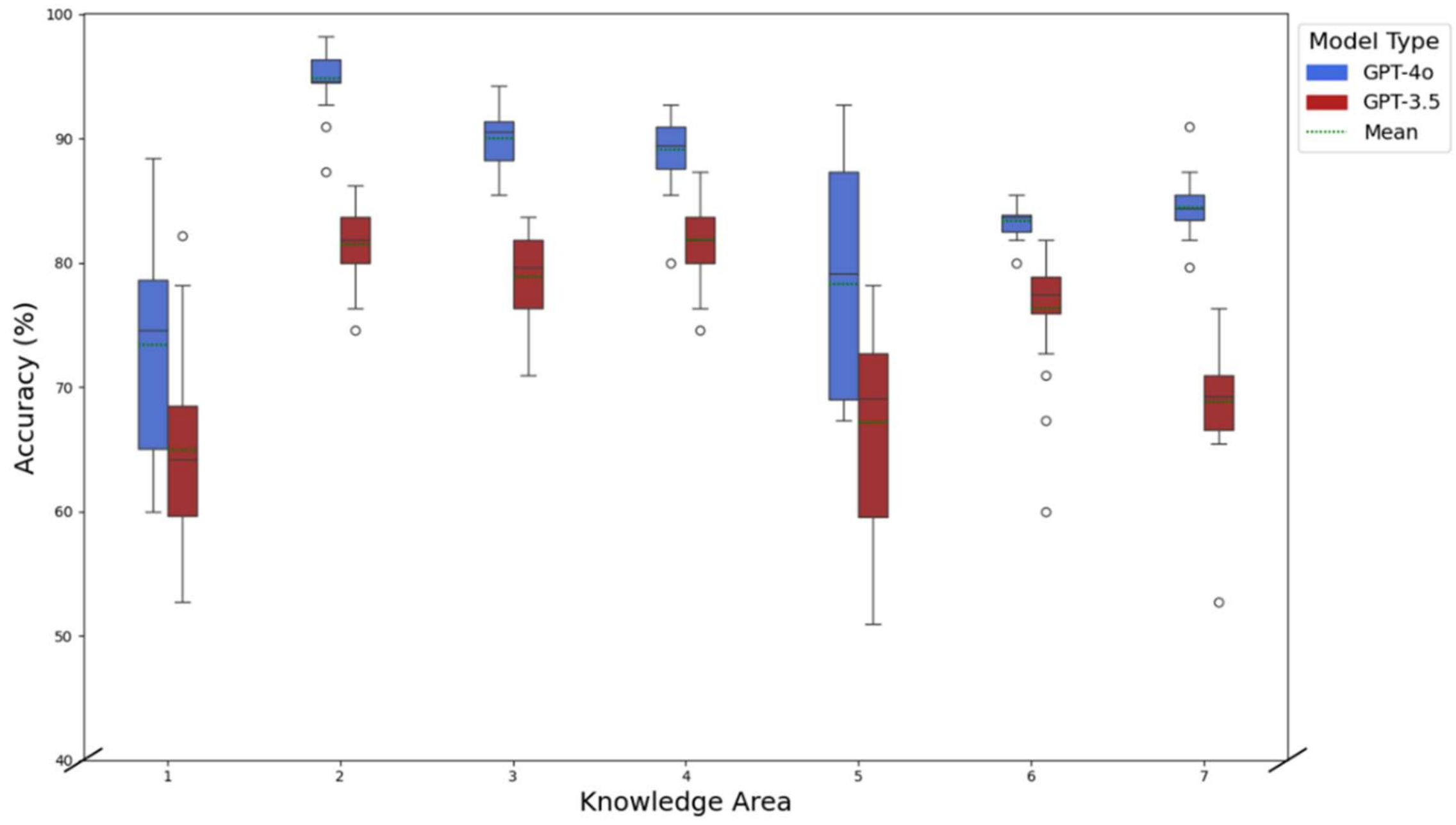

Figure 3: Accuracy comparison of GPT-3.5 and GPT-4o across KAs

Additionally, GPT models exhibited varying levels of knowledge across different KAs. The highest-performing subfields included Management Systems (KA 2), Ergonomics, Industrial Hygiene, and Occupational Health (KA 4), and Safety Hazard Identification and Control (KA 3). In contrast, the subfields where the models faced the greatest challenges were Science and Math (KA 1) and Emergency Response and Fire Prevention (KA 5). These differences between the two models and across the seven KAs are confirmed by the overall ANOVA results (Table 2), which included all experimental factors. This variation in performance underscores the inherent strengths and limitations of LLMs in various areas of safety expertise, which may be attributed to the availability and quality of pre-training data.

The overall ANOVA results (Table 2) also revealed that none of the factors related to prompting techniques and output requirements significantly affect accuracy. Despite this initial finding, we observed that the effectiveness of a prompting technique or output format can vary significantly depending on the specific model and KAs examined. Table 3 illustrates this variability by presenting the top five best-performing and bottom five lowest-performing configurations for both GPT-3.5 and GPT-4o models. A striking observation is that the second-best configuration for GPT-4o ($T_{FS}O_{A_1st_Spelled_Inc}$) was, in fact, the worst-performing configuration for GPT-3.5. Additionally, for GPT-4o, the best and worst configurations differ only in reasoning order, with the top-performing configuration placing reasoning first. This heterogeneity potentially results in cancelling effects in the aggregate analysis, as positive effects in one subset were offset by negative effects in another.

To further investigate potential effects that may have been masked in the aggregate analysis, separate ANOVA analyses were conducted for each KA-model combination. Table 4 presents the statistical results, while Figure 4 provides a visual representation with green dotted boxes highlighting the factors that demonstrated a statistically significant impact on performance. The results confirmed that no single prompt configuration is universally optimal across all KAs. An example is KA 1 (Science and Math), where significant effects were found for both reasoning order and reasoning inclusion for both GPT-3.5 and GPT-4o. Uniquely, only this KA showed significant improvement from including reasoning in the output and presenting it before the answer. This preference likely stems from the distinct nature of scientific and mathematical problems, which often require step-by-step problem-solving approaches. Table 4 and Figure 4 also revealed model-specific preferences: GPT-3.5's accuracy was significantly influenced by answer format across five subfields (KA 2, KA 3, KA 4, KA 6, and KA 7), where it consistently favored the ABCD format, with the exception of KA 5 favoring a spelled-out format. Meanwhile, GPT-4o showed a clear preference for the spelled-out format in KAs where output format had significant effects (KA 1, KA 3, and KA 5). Altogether, these findings highlight that there is no one-size-fits-all approach to prompting engineering and output requirements. The distinct preferences observed across KAs and between models underscore the need for subfield-specific and model-specific optimization to maximize LLM performance in safety-related tasks.

**Table 4:** ANOVA Results by Knowledge Area for GPT-3.5 and GPT-4o

| | GPT-3.5 | | | | GPT-4o | | | |
|---|---|---|---|---|---|---|---|---|
| **KA** | **Reasoning Order** | **Prompt Technique** | **Answer Format** | **Reasoning Inclusion** | **Reasoning Order** | **Prompt Technique** | **Answer Format** | **Reasoning Inclusion** |
| **1** | **0.029 *** | 0.858 | 0.458 | **0.007 **** | **0.039 *** | 0.540 | **0.021 *** | **0.013 *** |
| **2** | 0.103 | **0.009 *** | **0.005 **** | 0.303 | 0.532 | 0.159 | 0.149 | 0.668 |
| **3** | 0.796 | 0.620 | **0.012 *** | 0.513 | 0.86 | 0.503 | **0.027 *** | **0.049 *** |
| **4** | 0.316 | 0.705 | **0.001**** | 0.223 | 0.659 | 0.079 | 0.101 | 0.113 |
| **5** | 0.537 | 0.134 | **0.000 **** | 0.239 | 0.816 | 0.423 | **0.000 **** | 0.979 |
| **6** | 0.879 | 0.374 | **0.002 **** | 0.081 | 0.511 | 0.512 | 0.235 | 0.197 |
| **7** | 0.380 | 0.944 | 0.153 | 0.523 | **0.030 *** | 0.154 | 0.261 | 0.292 |

Note: Any statistically significant factor ($p < 0.05$) is bolded

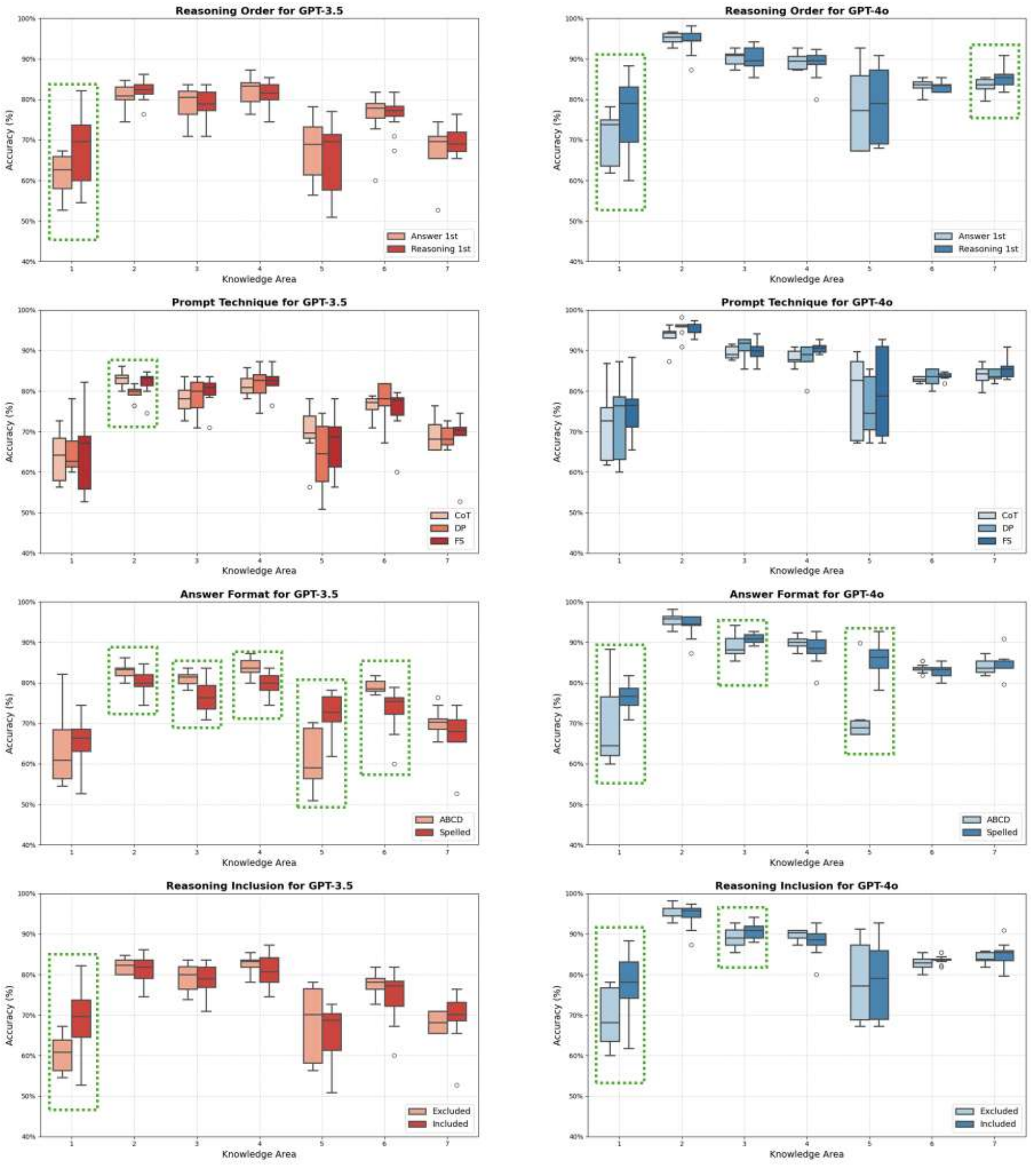

**Figure 4:** Comparison of Prompt Engineering Factors Across KAs for GPT-3.5 and GPT-4o. Note: Green boxes indicate statistically significant factors ($p < 0.05$).

**Error Analysis**

An error analysis of 42 outputs of the lowest-performing questions was conducted. Four types of error emerged, revealing the limitations of LLMs in areas where they encountered the greatest challenges: Lack of Knowledge (38%), Reasoning Flaws (31%), Memory Issues (24%), and Calculation Errors (7%).

*Lack of Knowledge*: LLMs may provide incorrect answers due to an inability to recall requisite information, such as the activation temperature of a red-colored fire sprinkler head. In such instances, models often resort to over-generalization or guesswork based on limited available information. More concerning are instances of hallucination, where the model fabricates information, such as inventing non-existent OSHA standards. Interestingly, this knowledge gap manifests inconsistently when LLMs are prompted with the same questions multiple times: The models may demonstrate inadequate knowledge in one attempt yet display sufficient knowledge in another.

*Reasoning Flaws*: Even when equipped with adequate knowledge, LLMs may misapply it due to incomplete or flawed reasoning processes. A notable example involved a free fall distance calculation where the anchor point is above the attachment point. In that case, the model correctly recalled relevant terms but proceeded directly to an incorrect answer without proper reasoning. Consequently, it failed to consider the specific scenario provided and defaulted to a common situation answer (i.e., the anchor point and the attachment point are at the same height) that was not applicable. In a separate response to the same question, the model did consider the context of the anchor point's position but arrived at an erroneous conclusion due to logical errors in its process.

*Memory Issues*: LLMs may exhibit difficulty retaining details from user inputs, particularly during multi-step processes and when managing a wider context. For example, after successfully calculating the NIOSH lifting index for a given scenario with detailed and accurate computations, the model inexplicably believed that the correct answer it had just calculated was not among the listed options. This suggests that the lengthy calculation process may have interfered with the LLM's ability to retain the original answer choices. A recency bias was also observed within the *Memory Issues* category, where LLMs favored or were entirely misled by recent information at the expense of earlier context. In one example involving crane inspection best practices, the first two options focused on record retention duration, while the latter two addressed responsible personnel. After analyzing the initial options, the model appeared to fixate on the record retention aspect, disregarding the overarching question of "best practices." This led to the incorrect dismissal of the final two options simply because they did not pertain to duration.

*Calculation Errors*: LLMs occasionally made arithmetic errors after correctly setting up the problem. For example, recurring mistakes in operations involving powers, such as calculating 15 raised to a power, revealed a weakness in the model's numerical execution capabilities.

***Experiment 2: Reliability and Consistency***

Reliability was evaluated through repeated measures ANOVA. The analysis found no significant differences in average accuracy across repetitions for any configuration tested ($p = 0.957$). These results indicate that LLMs consistently deliver reproducible performance in safety standard exams when model parameters are tightly controlled. The study further explored the consistency of LLMs using Entropy values, ranging from 0 to 2.32 bits. These values were calculated for each question and then averaged, standardized, and grouped into 10% accuracy bins for further comparison. As shown in Figure 5, there was a high consistency in the model's responses, with average entropy remaining low across all accuracy ranges. The highest observed entropy was 0.31 bits (13.2%), occurring in the 21-30% accuracy range, indicating that even when LLMs face more difficult questions, their consistency remained robust. Notably, 52.2% of the questions (201 out of 385) within the 91-100% accuracy range display exceptionally low entropy (1.1%), reflecting the model's strong consistency for questions it understood well.

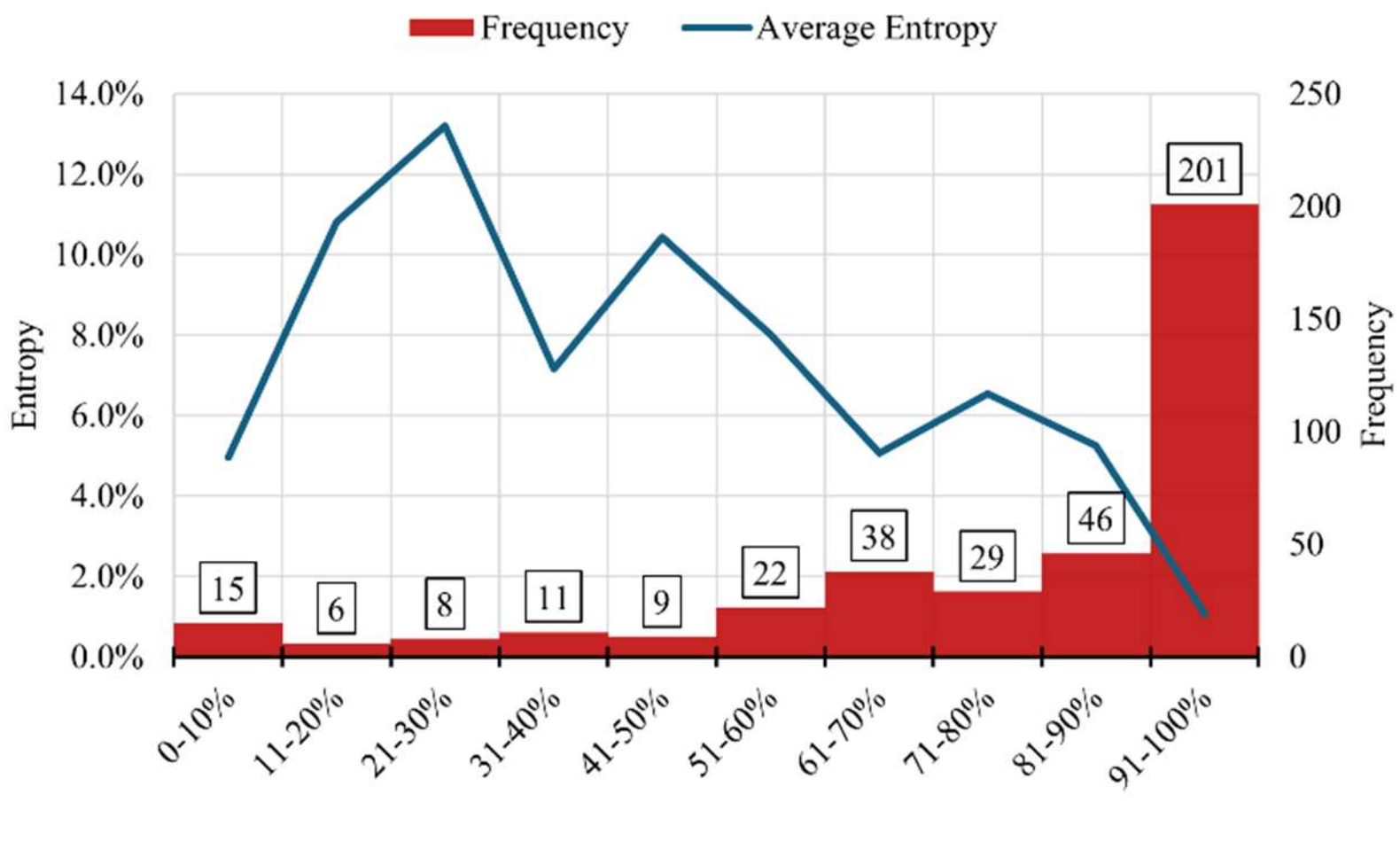

Figure 5 : Entropy and Frequency Distribution by Accuracy Bins

## DISCUSSION

### *Competency in Construction Safety*

This study offers a pioneering assessment of LLMs' subfield-specific knowledge within construction safety using standardized BCSP exams. We found compelling evidence that GPT-3.5 and GPT-4o have demonstrated proficiency in safety knowledge with test scores of 73.81% and 84.64%, respectively. Both scores exceed the typical passing threshold of approximately 60%. The reliability of these results was further validated through multiple exam attempts. These findings highlight the promising potential of LLMs within construction safety practices, particularly in the areas GPT-3.5 and GPT-4o exhibit notable proficiency, such as management system and health and safety hazard identification and control. BCSP exam questions in these areas assess safety professional's understanding of safety and health concepts, along with their ability to analyze specific scenarios, identify major hazards, and recommend appropriate control measures. LLMs' competence in these areas opens the door for numerous immediate applications. For instance, they can be leveraged to explain complex safety concepts in layman's terms and provide rapid second opinions on hazard assessments and control strategies. Furthermore, the proficiency of LLMs in safety-related concepts and terminology positions them to handle routine administrative tasks, such as extracting relevant information from diverse documents and automatically populating safety forms. This capability provides a distinct advantage over classical natural language processing (NLP), which often struggles with specialized vocabulary and contextual meaning (Nouhaila et al. 2024).

***Common Errors and Countermeasures***

While LLMs show considerable promise, it is crucial to acknowledge their common errors and challenges. This study identified a notable deficiency in pre-trained knowledge within certain subfields, such as emergency response and fire prevention. Additionally, our analysis uncovered specific “red flags” in GPT-generated responses that indicate a lack of requisite knowledge. These red flags include overgeneralization without considering context-dependent nuances and reliance on guesswork using elimination-based strategies. To mitigate the risk of receiving misleading or fabricated information, construction professionals can instruct the model to decline to answer questions outside its knowledge scope or request confidence scores (Yang et al. 2023). These approaches align with responsible AI practices, ensuring that LLMs are used in ways that prioritize safety and minimize the risk of harmful outcomes. Although these prompting techniques offer proven incremental benefits, Retrieval-Augmented Generation (RAG) presents a long-term solution to eliminate knowledge gaps by enabling LLMs to access external information sources and ground their responses in factual data (Krishna et al. 2024). Nevertheless, RAG is not without its limitations. LLMs may inconsistently prioritize retrieved information over their pre-trained knowledge, creating a "tug-of-war" effect (Alghisi et al. 2024). Furthermore, the "lost-in-the-middle" problem, where LLMs struggle to access information from the middle of large document sets (Hsieh et al. 2024), demands further investigation. Therefore, additional research is necessary to ensure RAG can be applied reliably within the construction safety context.

Beyond accessing factual information, LLMs must demonstrate the ability to effectively apply retrieved knowledge in context. However, this study observed that across various KAs, LLMs occasionally draw incorrect conclusions with logical leaps and exhibit flawed reasoning. These reasoning errors could be mitigated using clearer prompts to better guide the model's reasoning paths. In addition to COT and FS prompting techniques tested in this study, future research could explore approaches like Iteration of Thought (IOT) (Radha et al. 2024), which encourages the exploration of multiple reasoning paths and may yield more accurate responses. In other instances, LLMs could struggle to retain user-provided details during extended reasoning processes and show a bias toward recent information. This issue likely stems from limited context window size and the absence of memory structures. Although these computational constraints may be overcome with rapid advancements in generative AI technology, some interim strategies have proven effective in enhancing LLMs' memory management. These include prompting the model to reread the question (Xu et al. 2023), structuring user input clearly using placeholders to highlight key information (He et al. 2024), removing irrelevant details from prompts (Shi et al. 2023), and breaking down complex queries into sequential sub-questions.

A more sophisticated, automated solution to address reasoning flaws and memory limitations is the implementation of an agentic workflow (Wu et al. 2023). This approach leverages multiple LLM-enabled agents, each assigned a specific task—such as decomposing questions, answering sub-questions, reviewing reasoning soundness, and synthesizing results. By distributing these responsibilities, each agent can specialize in its designated function, thus reducing the cognitive load and memory pressure on any single model. Fine-tuning offers another advanced solution, wherein a pre-trained model undergoes further training on task-specific data. This adaptation process enables the model to assimilate specialized knowledge that may be absent from the original model and refines its behavior, such as self-correction and attention allocation habits. However, fine-tuning requires careful consideration. It is primarily used to tailor models to specific tasks, and while this can enhance performance in the targeted area, it may lead to reduced effectiveness outside that specialization (Wang et al. 2022). Fine-tuning may also inadvertently weaken the security guardrails embedded in the original models, compromising their ethical use (Wang et al. 2022). Therefore, it is crucial to judiciously determine when and how to apply fine-tuning.

Notably, GPT-3.5 and GPT-4o encountered significant challenges in science and mathematics, domains that typically involve numerical operations. Our observations aligned with the findings of Bharatha et al. (2024) and Amirizaniani et al. (2024), revealed that LLMs are susceptible to calculation errors. In the context of construction safety, such errors can have dire consequences, such as fatal falls resulting from incorrect clearance calculations. Given that LLMs are primarily designed for natural language processing rather than numerical computations, construction professionals should exercise caution when relying on these models for any form of calculation. Future research could explore the integration of specialized mathematical tools or machine learning algorithms into LLMs via function calling through application programming interfaces (APIs) (Yuan et al. 2024). Such integration could yield a more comprehensive technological solution capable of addressing complex safety problems that require a seamless combination of numerical computations and natural language generation.

***Lack of Interpretability in LLMs***

This study uncovered several phenomena that underscore the challenge of interpretability in LLMs. We observed considerable variability in the effectiveness of different prompting techniques and output requirements across various KAs and LLMs. While certain model behaviors—such as improved performance on mathematical tasks when explicit reasoning is included—are explicable, many others remain unknown. For instance, we still do not fully grasp why certain combinations of prompts and output formats yield excellent results in one model but prove ineffective in another (Table 3**Error! Reference source not found.**). Moreover, the inherent probabilistic nature of LLMs complicates interpretability efforts (Callanan et al. 2023). Even when external factors, such as model temperature, seed number, and context window, were controlled, we encountered instances where identical queries resulted in different responses. This variability makes it difficult to trace back the reasons behind a specific output, thereby hindering efforts to fully trust the model.

Interestingly, our study revealed human-like biases in LLMs when applied to construction safety problem-solving. These biases, including recency bias, closely resemble cognitive biases observed in human decision-making, although the underlying mechanisms may differ (Yax et al. 2024). Similar to human behavior, the actions of LLMs often prove difficult to predict. These findings highlight the urgent need for further research to clarify the internal mechanisms driving LLM behavior. As a result, this study does not advocate for a single, definitive combination of prompt techniques and output formats. Our findings suggest that a universally applicable approach is improbable. Instead, a context-specific strategy of trial and error, where prompts are carefully selected and adapted based on the task and the specific LLM being used, remains the most effective approach.

***Research Limitations and Future Research***

This study has several limitations that warrant cautious interpretation of the results and suggest avenues for future research. First, the BCSP is a leading organization in the U.S. for occupational safety and health certification, and their exams were selected in this study as an authoritative assessment of a broad range of domain knowledge. However, they may not sufficiently evaluate specialized expertise in certain niche areas. Future studies should incorporate more specialized assessments to better gauge LLMs' proficiency within that specific domain. Second, the reliance on MCQs, while providing objective evaluation, limits the assessment of LLMs' ability to creatively and meaningfully apply knowledge in real-world scenarios. For instance, while LLMs' medical knowledge has been found comparable to that of clinical experts, their decision-making performance across various stages of the clinical diagnostic process has been shown to lag behind (Hager et al. 2024). This study has analyzed LLMs' reasoning outputs and provided error analysis in the context of safety. Subsequent research should expand the scope of inquiry by incorporating long-form, open-ended questions that require contextual application of knowledge. Furthermore, it is crucial to evaluate LLMs' command of safety terminology and knowledge in languages other than English, as this study only tested them in an English-language context. Future research should assess their performance across multiple languages to determine their effectiveness for multilingual construction safety training and communication. Lastly, given the rapid evolution of LLM capabilities, these findings represent a snapshot in time and may not reflect the most current state of the technology. Continuous monitoring is needed to keep pace with advancements in the field.

## CONCLUSIONS

The integration of LLMs into construction engineering and management offers both opportunities and challenges. This study pioneers a systematic evaluation of LLMs' performance in construction safety and makes three significant contributions to the field. First, through standardized testing, we establish a baseline understanding of LLMs' safety knowledge, demonstrating their potential as virtual experts in tasks such as hazard identification, the development of control measures, and the structuring of management systems. This capability is especially beneficial for smaller contractors, who often lack dedicated safety personnel and are disproportionately represented in industry fatality statistics. Larger construction firms can also leverage LLMs to automate safety-related administrative functions and streamline management processes. However, our study identifies key knowledge gaps in areas such as science and mathematics, emergency response, and fire prevention. These findings underscore the necessity of human oversight in LLM deployment and help pinpoint where such oversight is most critical.

Second, our findings indicate that the successful implementation of LLMs relies on effective prompt engineering and output specification. This study is the first to illustrate that LLM effectiveness can vary substantially across different models and types of safety knowledge areas. This insight underscores the need for construction professionals to adopt tailored, iterative approaches to prompt development instead of relying on one-size-fits-all solutions. Third, we highlight several intrinsic limitations of current LLMs in construction safety applications. These include a propensity for calculation errors, logical inconsistencies, and difficulty in retaining user-provided information during long or complex interactions. Such limitations, revealed through our empirical analysis, highlight the need to enhance LLM capabilities through advanced techniques like model fine-tuning, RAG, and agentic workflow engineering.

In conclusion, this research provides a critical foundation for the responsible integration of LLMs into construction safety practices. By addressing the identified limitations and pursuing the suggested avenues for future research, we can unlock the full potential of these technologies to create safer working environments and ultimately move closer to the vision of zero injuries in construction.

**DATA AVAILABILITY STATEMENT:**

Data that support the findings of this study are available from the corresponding author upon reasonable request.